\documentclass[11pt]{article}

\usepackage[T1]{fontenc}
\usepackage[utf8]{inputenc}
\usepackage{mathptmx}
\usepackage[a4paper,margin=2.5cm]{geometry}
\usepackage{setspace}
\usepackage{amsmath,amssymb}
\usepackage{graphicx}
\usepackage{booktabs}
\usepackage{tabularx}
\usepackage{array}
\usepackage[labelfont=bf,labelsep=space,font=small]{caption}
\usepackage[numbers,sort&compress]{natbib}
\usepackage{authblk}
\usepackage{xcolor}
\usepackage{lineno}
\usepackage{microtype}
\usepackage{seqsplit}
\usepackage[hidelinks]{hyperref}

\newcolumntype{L}{>{\raggedright\arraybackslash}X}

\title{Testing concept-level audits of melanoma classifiers with diffusion counterfactuals and matched positive controls}

\author[1]{Md~Shohanoor~Rahman}
\author[2]{Md~Hasibul~Islam}
\author[3]{Md~Kishor~Morol}
\author[2]{Dip~Nandi}
\author[4,*]{Tze~Hui~Liew}
\affil[1]{Department of Computer Science \& Engineering, Rajshahi University of Engineering \& Technology, Rajshahi, 6204, Bangladesh}
\affil[2]{Department of Computer Science \& Engineering, American International University-Bangladesh (AIUB), Dhaka, Bangladesh}
\affil[3]{ELITE Research Lab, Queens, NY, 11435, USA}
\affil[4]{Centre for Intelligent Cloud Computing (CICC), COE of Advanced Cloud, Faculty of Information Science \& Technology, Multimedia University, Jalan Ayer Keroh Lama, Bukit Beruang, Melaka, 75450, Malaysia}
\affil[*]{Corresponding author: Tze~Hui~Liew (\href{mailto:thliew@mmu.edu.my}{thliew@mmu.edu.my})}
\date{}

\begin{document}
\maketitle

\begin{abstract}
\noindent
Concept-based explanations of melanoma classifiers are rarely tested against controls with known targets. We built a counterfactual audit from a diffusion model conditioned on seven 7-point checklist concepts and tested it with such controls. Colour-calibrated start noise reduced colour outliers from 78.5-84.8\% to 6.6-13.3\%. Four of seven knobs were target-dominant but also changed other concepts. Quartile-based positive controls were confounded, and the audit failed its pre-specified rule. It recovered the targets of all 16 controls matched on the generator's pseudo-labels and of 12 of 12 controls labelled by a second evaluator, for three concepts; concept activation vectors recovered fewer (11 of 16). In 18 permuted-label classifiers, significant sweep responses and target rankings also arose, whereas standardised effect margins separated the controls from these null classifiers. The pigmentation sweep raised the melanoma logit in 15 of 15 ISIC 2020 classifiers (+3.12 on average) and 6 of 6 HAM10000 classifiers (+4.01), about four times the null response in scale-free units; the regression effect did not transfer. An exploratory decomposition, evaluated against the controls, suggested that knob cross-talk partly hides vascular responses. Matched controls and null classifiers are practical checks of concept audits.
\end{abstract}

\section*{Introduction}

A melanoma classifier can reach a high AUC for the wrong reasons. Surgical skin markings raised the melanoma scores of a certified convolutional network and lowered its specificity from 84.1\% to 45.8\% \cite{winkler2019markings}. A study of seven visual artefacts concluded that current debiasing methods do not yet remove this kind of bias from skin-lesion models \cite{bissoto2020debiasing}. DeGrave et al.\ combined generative counterfactual images with the judgement of dermatologists to audit five dermatology classifiers. The classifiers used lesional pigmentation patterns, as dermatologists do, but they also used background skin texture and colour balance \cite{degrave2025auditing}. A clinician who reads a model explanation therefore needs to know whether the explanation itself is correct.

Concept-based methods explain a classifier in clinical terms. Testing with concept activation vectors (TCAV) measures how sensitive a prediction is to a direction in activation space that separates images with a concept from random images \cite{kim2018tcav}. In dermoscopy, TCAV has been used to test whether skin-lesion classifiers rely on dermoscopic concepts \cite{lucieri2020cav,lucieri2022exaid}. The 7-point checklist scores seven dermoscopic structures \cite{argenziano1998sevenpoint}, and the Derm7pt dataset gives expert labels for them \cite{kawahara2019derm7pt}. Such explanations, however, are rarely tested against a reference with a known target. Some saliency maps do not change when the model weights are randomised \cite{adebayo2018sanity}, and concept explanations change with the probe dataset and with how well a concept can be learned \cite{ramaswamy2023overlooked}. Known-answer tests have so far relied on planted artificial signals. Adebayo et al.\ found that common explanation methods did not reliably reveal a planted spurious signal \cite{adebayo2022posthoc}, and Pahde et al.\ compared CAVs with ground-truth directions in controlled data and in ISIC 2019 classifiers with inserted shortcuts \cite{pahde2025navigating}. Such tests do not show whether a concept method recovers the clinical dermoscopic structures that a classifier actually uses. TCAV and regression on concept scores also share a causal limitation: they measure association, and when concepts occur together, an association with one concept can stand for another \cite{goyal2019cace,pearl2009causality}. Goyal et al.\ therefore proposed to estimate the causal effect of a concept with a generative model \cite{goyal2019cace}, and CEBaB provides counterfactual texts as a benchmark for such estimators in language models \cite{abraham2022cebab}.

A generative model makes this kind of intervention possible in images. In dermatology, Varshney et al.\ derived concepts from latent-diffusion counterfactuals of an ISIC classifier \cite{varshney2024counterfactual}, MONET scores images for clinical concepts with an image-text foundation model in order to audit data and models \cite{kim2024monet}, and concept-guided diffusion has been used to synthesise training images \cite{fayyad2025lesiongen}. If a model can add or remove one dermoscopic structure while it keeps the rest of the image, the change in the classifier output is an interventional effect, provided that each knob changes the intended concept and little else. We trained a denoising diffusion probabilistic model \cite{ho2020ddpm} with classifier-free guidance \cite{ho2022cfg} on ISIC 2020 \cite{rotemberg2021isic}, conditioned on concept pseudo-labels from a concept bottleneck model \cite{koh2020cbm}. Such an instrument has its own failure modes: a knob can move several concepts, the images can be unrealistic, and the colour can drift. An audit made with it is only as good as the evidence that it finds what a classifier really uses.

\begin{figure}[tbp]
\centering
\includegraphics[width=\textwidth]{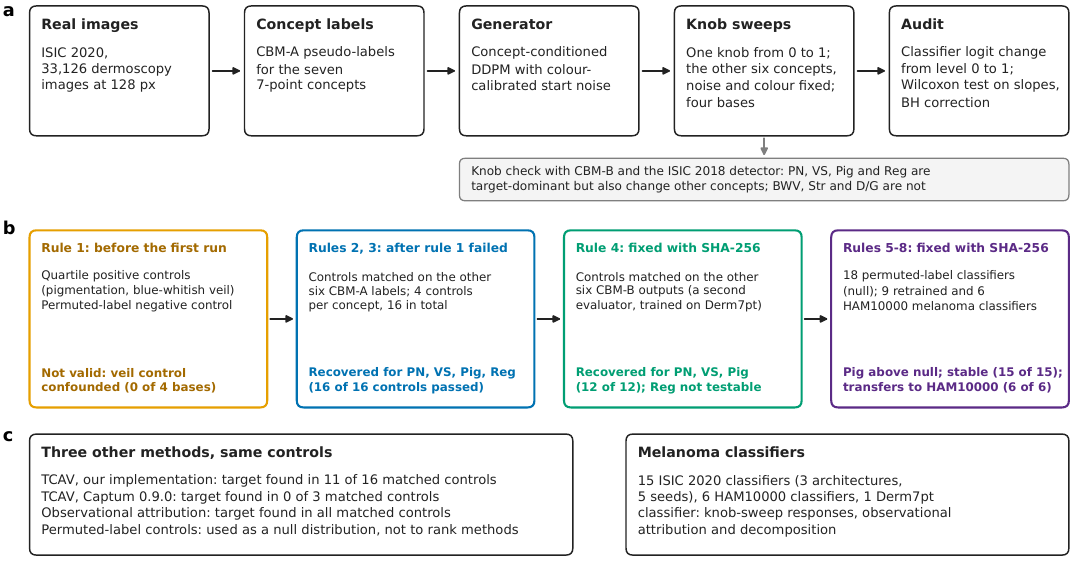}
\caption{\textbf{Study design.} \textbf{a}, The counterfactual instrument and the audit statistic. CBM-A, the concept bottleneck model that produced the conditioning pseudo-labels; CBM-B, an evaluator trained separately on the Derm7pt expert labels. \textbf{b}, The validity rules in the order in which they were fixed, with their outcomes. Each rule was written into the code before the run that tested it. \textbf{c}, The three other methods, tested on the same controls, and the melanoma classifiers. PN, pigment network; VS, vascular structures; Pig, pigmentation; Reg, regression structures; BWV, blue-whitish veil; Str, streaks; D/G, dots and globules; BH, Benjamini-Hochberg.}
\label{fig:design}
\end{figure}

To our knowledge, no study has tested concept explanations of dermatology classifiers against control classifiers trained on known targets for clinical dermoscopic concepts, or compared TCAV with an interventional audit under the same controls. We therefore asked whether a concept-level counterfactual audit, TCAV and observational attribution recover the training targets of such controls: classifiers trained to detect one concept, with a permuted-label classifier as a negative control. Each decision rule was fixed in code before the run that tested it, and we report the rule that failed as well as the rules that passed. We show that quartile-based positive controls are confounded, that the audit recovers the targets of matched controls more often than TCAV, and that 18 permuted-label classifiers, used as a null distribution, also show significant sweep responses and target rankings. Against this null, the response of melanoma classifiers to the pigmentation sweep holds across retraining and a second training dataset. Figure~\ref{fig:design} summarises the study design.

\section*{Results}

\subsection*{A colour-calibrated generator with four target-dominant knobs}

The instrument is a concept-conditioned DDPM trained on ISIC 2020 images at $128\times128$ pixels. Its conditioning vector holds the CBM-A pseudo-labels of the seven 7-point checklist concepts (Methods). Because CBM-A produced these labels, two other evaluators judged the generated images: CBM-B, trained separately on the Derm7pt expert labels \cite{kawahara2019derm7pt} (mean concept AUC 0.849 at 128 pixels), and an ISIC 2018 Task 2 attribute detector \cite{codella2019isic2018}, which gave a second reading for pigment network (AUC 0.786). On 200 PH2 images \cite{mendonca2013ph2} with expert criteria from another centre (rule 8), CBM-B reached AUCs of 0.84 to 0.92 for streaks, regression and veil, 0.70 for atypical dots and globules, and 0.58 (0.51 to 0.66) for atypical against typical pigment network; PH2 does not score vascular structures or pigmentation.

\begin{figure}[tbp]
\centering
\includegraphics[width=\textwidth]{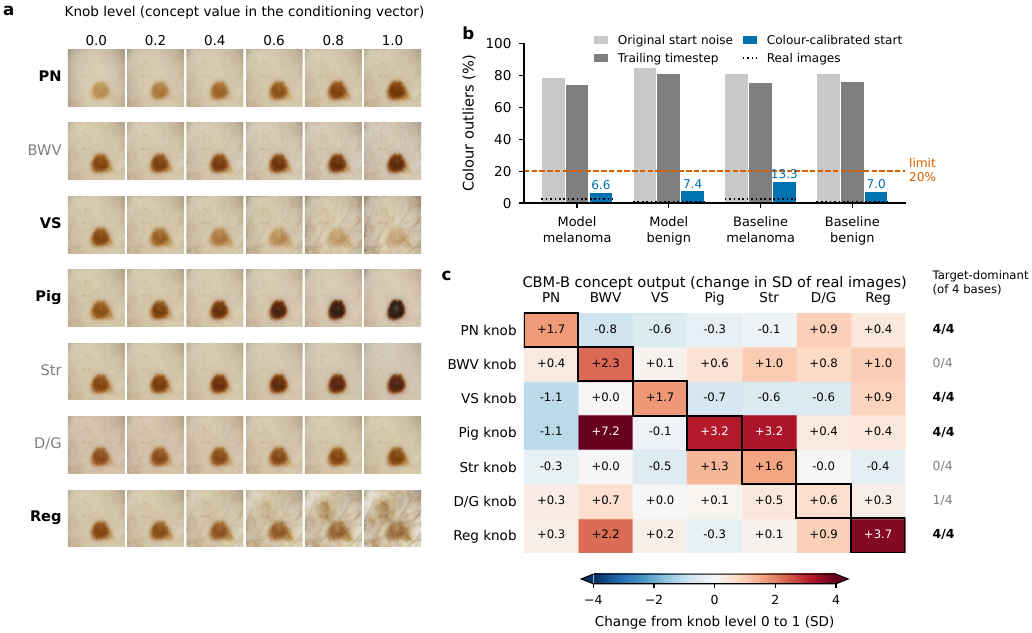}
\caption{\textbf{The counterfactual instrument.} \textbf{a}, AI-generated images (our concept-conditioned diffusion model; research outputs of this study, not clinical photographs) for one realistic benign base (the pseudo-label vector of a real ISIC 2020 benign image) with each of the seven concept knobs set from 0 to 1. Along each row the other six concepts, the start noise and the colour target do not change. Bold labels mark the four target-dominant knobs; grey labels mark the other three. \textbf{b}, Colour outliers (images whose mean CIELAB lies outside the 99\% region of real images) for the original start noise, a trailing-timestep start and the colour-calibrated start, with the same 256 start noises per set. Dotted lines, real images; dashed line, the 20\% limit fixed before the test. \textbf{c}, Mean change of each CBM-B concept output, in SD of real test images, when each knob goes from 0 to 1 (mean of the two realistic bases, 64 trajectories each). Outlined cells, the knob's own concept. Right, the number of the four bases in which the knob raised its own output more than any other knob did (target-dominant).}
\label{fig:instrument}
\end{figure}

Before any audit, the sampler needed a correction. In a paired test with 256 start noises per set, 78.5\% to 84.8\% of the prototype images of the two models fell outside the 99\% colour region of real images with the original start noise, against 2.4\% and 0.6\% for real melanoma and benign images. A trailing-timestep start \cite{lin2024schedules} did not solve the problem. The source was the start noise itself: its channel means predicted the mean colour of the final image (cross-validated $R^2 = 0.90$), an instance of the signal-leak bias described by Everaert et al.\ \cite{everaert2024signalleak}. We therefore set these means so that each image reaches a colour target drawn from the real colour distribution of its class (Methods). With the same start noises, the outlier rate fell to 6.6\% to 13.3\% (Fig.~\ref{fig:instrument}b), and to 2.3\% to 8.9\% on the final evaluation sets. The targets did not carry the class signal on their own: with one class-agnostic target distribution, CBM-B still separated melanoma from benign prototype images (AUC 0.947, against 0.966 with class-specific targets).

A concept knob is the value of one concept in the conditioning vector. To test a knob, we moved it from 0 to 1 while the other six concepts kept the values of a base vector. We used four bases: the pseudo-label vectors of 64 benign and 64 melanoma training images, which we call the realistic bases, and the benign and melanoma class prototypes. We call a knob target-dominant when it raised its own CBM-B output more than any other knob did. The pigment network, vascular structures, pigmentation and regression knobs were target-dominant in all four bases (Fig.~\ref{fig:instrument}c), and the ISIC 2018 detector confirmed the pigment-network knob in all four. Target dominance does not mean that a knob changes one concept only. On the realistic bases, the pigmentation knob raised the CBM-B veil output (+7.2 SD) more than its own (+3.2 SD) and raised streaks by 3.2 SD, the regression knob raised veil by 2.2 SD, and the vascular knob lowered pigment network by 1.1 SD. The veil, streaks and dots-and-globules knobs were not target-dominant. An audit effect is therefore the response of a classifier to the whole sweep of a knob, not to a single concept.

The class prototypes produced less realistic images. Inception KID ($\times 10^{3}$) to real melanomas was 139.7 for the melanoma prototype, against 35.2 for the class-conditioned baseline model and 27.3 for images generated from real melanoma vectors. No melanoma-conditioned image was closer to a real training melanoma than the 99th percentile of real-to-real nearest-neighbour similarity. The realistic bases therefore carry the main analysis, and the prototypes enter only the validity rules.

\subsection*{Classifiers under audit}

We audited 26 binary classifiers (Supplementary Table S1), and a later run added 12 controls labelled by CBM-B (see below). Six melanoma classifiers were trained on ISIC 2020, with ResNet-50, EfficientNet-B0 and ViT-B/16 backbones and two seeds each (test AUC 0.870 to 0.913). A seventh, a ResNet-50, was trained on Derm7pt (test AUC 0.833 on Derm7pt and 0.673 on ISIC 2020). The other 19 classifiers were controls with known training targets: two quartile positive controls for pigmentation and blue-whitish veil, a negative control trained on permuted melanoma labels, and 16 matched positive controls. A third run, with rules fixed before it (rules 5 to 8, Methods), added nine ISIC 2020 melanoma classifiers with new seeds (seeds 3 to 5), six classifiers trained on HAM10000 \cite{tschandl2018ham10000} and 17 permuted-label classifiers (Supplementary Table S25). Clipping of the logits at $\pm15$ affected up to 4.38\% of the generated images for the EfficientNet-B0 classifiers (Supplementary Table S2).

\subsection*{Quartile controls failed, and matched controls passed}

Before the first audit run, we fixed a validity rule in code (Methods): each positive control had to show its target as the largest positive and significant effect in at least 3 of 4 bases, and the largest absolute effect of the negative control had to stay below 0.25 times the median for the melanoma classifiers. The negative control met this condition (0.091 against a median of 3.99 logits; but see below), and the pigmentation control passed in 4 of 4 bases. The blue-whitish-veil control, however, failed in all four bases. By its own rule, the audit was not valid.

We traced the failure to the control's design. Blue-whitish-veil pseudo-labels are near zero for almost every image (median 0.0004; only 41 of 33,126 reach 0.5; Supplementary Table S3). As a result, the quartile groups differed by only 0.33 SD in the veil label, but by 1.22 SD in vascular structures and by $-1.12$ SD in pigment network (Fig.~\ref{fig:controls}a). The audit reported this pattern: the vascular and pigmentation knobs moved the control by +4.9 logits each, the veil knob, which is not target-dominant, by +1.2 (Fig.~\ref{fig:controls}c).

To reduce this confounding, we built matched positive controls for the four target-dominant concepts: for each, high-label and low-label training images were paired 1:1 by nearest-neighbour matching on the other six standardised pseudo-labels, with exact matching on the melanoma label (Methods). The pairs differed by 2.14 to 4.17 SD in the target label and by no more than 0.158 SD in any other label (Fig.~\ref{fig:controls}a). Each concept received four controls (ResNet-50 seeds 1 to 3, ViT-B/16 seed 1; test AUC for their own label 0.909 to 0.998). We fixed this design, and the rule that at least 3 of the 4 controls of a concept must pass, after the original audit had failed but before the runs that tested them.

All 16 matched controls passed (Fig.~\ref{fig:controls}c). Thirteen passed in all four bases; the three others were regression controls, which passed in three because the vascular knob gave a slightly larger effect in the real benign base. On the realistic bases, the target knob moved the control logit by +3.1 to +8.8, and the four controls of each concept gave closely similar profiles (mean Spearman $\rho$ 0.93 to 0.96). By the rule, the audit passed for all four concepts. Because the rule concerns only the target effect, we do not interpret the off-target effects, several of which were large and negative (Fig.~\ref{fig:controls}c,d).

\subsection*{Controls labelled by a second evaluator confirmed three of the four concepts}

\begin{figure}[tbp]
\centering
\includegraphics[width=\textwidth]{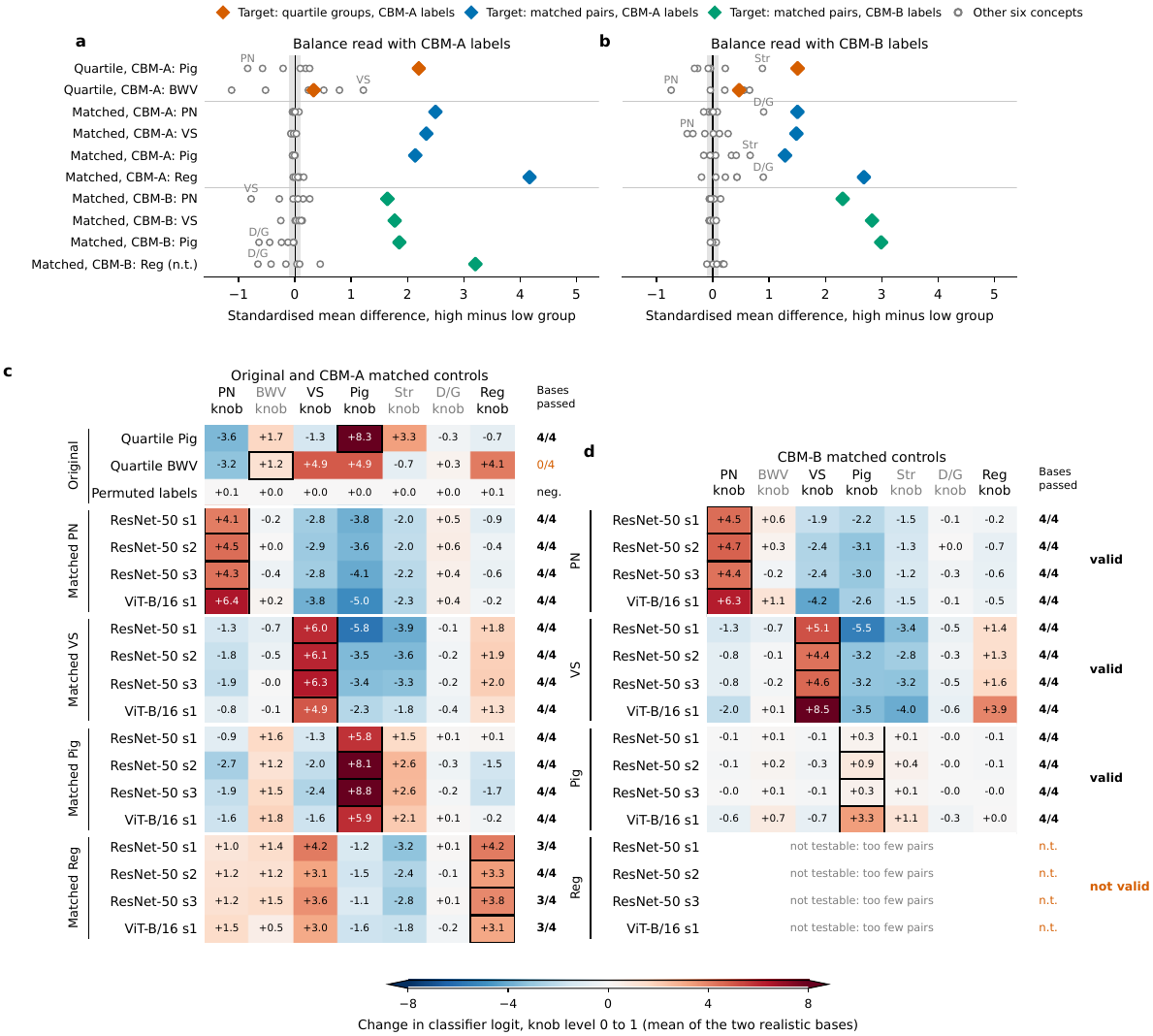}
\caption{\textbf{Positive controls of the audit.} \textbf{a}, \textbf{b}, Standardised mean difference between the high and low groups of each positive-control design, for the target concept (diamond) and the six other concepts (circles), read with the CBM-A pseudo-labels (\textbf{a}) and with the CBM-B outputs (\textbf{b}). Orange, quartile designs; blue, pairs matched on CBM-A labels; green, pairs matched on CBM-B labels. Each matched design is balanced in its own label set but not fully in the other. Labels name the most unbalanced other concept when its difference exceeds 0.3 SD. n.t., not testable (fewer than 50 training pairs). \textbf{c}, Change in the control classifier's logit from knob level 0 to 1 (mean of the two realistic bases) for the 3 original controls and the 16 CBM-A matched controls. \textbf{d}, The same for the CBM-B matched controls, with the decision of rule 4 for each concept (valid or not valid). Outlined cells, target knob. Right of each panel, the number of bases (of 4) in which the target knob gave the largest positive and significant effect; neg., negative control.}
\label{fig:controls}
\end{figure}

One weakness remained: the labels of these controls came from CBM-A, which also produced the generator's conditioning labels, so a shared CBM-A error could make a control and a knob agree without the concept itself. We therefore repeated the test with matched controls labelled by CBM-B. CBM-B was trained separately, on the same Derm7pt images, and its outputs on ISIC images are predictions, not expert annotations. Rule 4 (Methods) was printed with its SHA-256 fingerprint before the run computed any CBM-B label. Pigment network, vascular structures and pigmentation gave 154, 190 and 109 training pairs, differing by 2.31 to 2.99 SD in the CBM-B target output and by at most 0.140 SD in any other (Fig.~\ref{fig:controls}b; Supplementary Table S19). Regression gave only 37 training pairs, below the minimum of 50 set by the rule, so it was not testable and counts as not passed.

All 12 controls of the three testable concepts passed, each in all four bases (Fig.~\ref{fig:controls}d; Supplementary Table S20). Their test AUC for their own label was 0.899 to 1.000. On the realistic bases the target knob moved the control logit by +4.4 to +8.5 for pigment network and vascular structures; the three ResNet-50 pigmentation controls responded weakly (+0.3 to +0.9), but their logit SD was small, and their standardised margins (below) were 0.63 to 0.99. The four controls of a concept share its pairs and differ only in seed or architecture, so they are correlated replicates: the unit of validation is the concept, not the control. The audit therefore passed with CBM-B labels for pigment network, vascular structures and pigmentation; for regression, it passed with CBM-A labels only.

Each set of pairs was balanced only in the labels that built it: read with the other evaluator, other concepts differed by up to 0.90 SD (Fig.~\ref{fig:controls}a,b). Since the two control sets carry different residual differences and both passed, the result for these three concepts depends less on either label set.

\subsection*{TCAV recovered fewer control targets}

\begin{figure}[tbp]
\centering
\includegraphics[width=\textwidth]{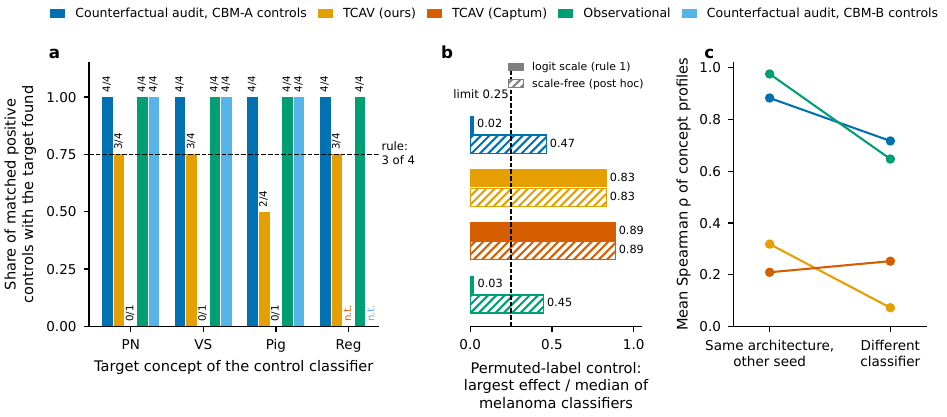}
\caption{\textbf{Four methods against the same controls.} \textbf{a}, Share of the matched positive controls in which each method found the target concept (counts above the bars). Light blue, the audit of the controls labelled by CBM-B; the other bars use the CBM-A controls. For the observational method, the target had the largest coefficient; it has no significance test. n.t., not tested. Dashed line, the per-concept rule (3 of 4). \textbf{b}, Largest effect of the permuted-label control divided by the median for the melanoma classifiers, on the logit scale of rule 1 (solid) and after dividing by each classifier's logit SD on the real test images (hatched, post hoc); TCAV scores do not depend on output scale. Dashed line, the limit of 0.25. \textbf{c}, Mean Spearman correlation of concept profiles between classifiers of the same architecture with another seed, and between different classifiers.}
\label{fig:methods}
\end{figure}

We then applied three other methods to the same classifiers (Fig.~\ref{fig:methods}, Table~\ref{tab:methods}): our TCAV implementation, the TCAV class of Captum \cite{kokhlikyan2020captum}, and an observational attribution. For TCAV, the rule fixed before the run required the target concept to have the highest mean TCAV score, significantly different from random CAVs, in at least 6 of 8 settings (Methods). We did not run TCAV on the CBM-B controls.

Our TCAV passed both quartile controls (6 of 8 settings each), including the veil control that the audit had rejected. This pass is circular, because the TCAV concept set and the control labels come from the same pseudo-label and share its confounds. On the matched controls, our TCAV passed 11 of 16. It failed all four ViT-B/16 controls and the seed-1 ResNet-50 pigmentation control, so it passed for pigment network, vascular structures and regression, but not for pigmentation (2 of 4). Its concept profiles agreed poorly between seeds and between classifiers (Table~\ref{tab:methods}).

Captum TCAV with its default settings passed each quartile control in 2 of 4 available settings, where 3 were needed, and none of the 3 matched controls tested. Of its TCAV scores, 95.8\% were exactly 0 or 1 (agreement with our implementation, mean $\rho$ 0.41). The coverage of the comparison is therefore unequal.

The observational attribution regresses the classifier logit on the CBM-B outputs and the mean CIELAB colour of real test images. Its largest coefficient was the target concept for all 16 CBM-A controls and all 12 CBM-B controls; the second test is circular, since the CBM-B outputs that define those labels are also its predictors. It gave the most stable profiles between seeds ($\rho$ 0.976). These passes have a limit: matching reduces the correlation between the target and the measured concepts, so matched controls cannot reveal the main weakness of an observational method, which appears when concepts are correlated, as in the data from which the melanoma classifiers learned.

The single permuted-label control of rule 1 did not separate the methods. On the logit scale, its ratio was 0.023 for the audit and 0.025 for observational attribution, against 0.834 and 0.889 for TCAV, but its logits varied little (SD 0.12 on the real test images, against 2.1 to 3.4 for the melanoma classifiers), and divided by each classifier's logit SD (post hoc) the audit and observational ratios were 0.47 and 0.45 (Fig.~\ref{fig:methods}b; Supplementary Table S22). The 17 later permuted-label classifiers had logit SDs from 0.05 to 1.97, and 11 of the 18 would have met the logit-scale condition of rule 1. We therefore use these classifiers as a null distribution (below), not to rank the methods; TCAV was not run on them.

\begin{table}[tbp]
\centering
\caption{\textbf{Validity and reliability of four concept-attribution methods on the same classifiers.} Negative-control ratio: largest absolute effect of the permuted-label classifier divided by the median of the same value for the melanoma classifiers, on the logit scale fixed in rule 1 and, in brackets, after dividing each classifier's effects by the SD of its logits on the real test images (post hoc); TCAV scores do not depend on output scale. $\rho$: mean Spearman correlation of concept profiles between classifier pairs. Captum TCAV was run on 13 classifiers, including the first three matched controls. CBM-B labels: 12 controls for pigment network, vascular structures and pigmentation; regression had too few pairs to test. For the observational method the CBM-B test is circular, because its predictors are the CBM-B outputs that define the control labels. PN, pigment network; VS, vascular structures; Pig, pigmentation; Reg, regression structures.}
\label{tab:methods}
\scriptsize
\setlength{\tabcolsep}{2.5pt}
\begin{tabularx}{\textwidth}{L L c L L c c c}
\toprule
Method & Pre-specified rule, quartile controls & \shortstack{Matched controls\\passed, CBM-A\\labels} & Matched controls passed, CBM-B labels & Concepts passing & \shortstack{Negative-\\control ratio\\(limit 0.25):\\logit (scale-free)} & \shortstack{$\rho$, same\\architecture,\\other seed} & \shortstack{$\rho$,\\different\\classifier} \\
\midrule
Counterfactual audit & Not valid: veil control 0 of 4 bases & 16 of 16 & 12 of 12 (Reg not testable) & PN, VS, Pig with both label sets; Reg with CBM-A labels only & 0.023 (0.47) & 0.883 & 0.717 \\
TCAV, our implementation & Not valid: negative control & 11 of 16 & Not tested & PN, VS, Reg with CBM-A labels; not Pig & 0.834 (0.834) & 0.318 & 0.072 \\
TCAV, Captum 0.9.0 & Not valid: both positives and the negative & 0 of 3 & Not tested & Not tested & 0.889 (0.889) & 0.209 & 0.252 \\
Observational attribution & No rule fixed in advance & 16 of 16 (largest coefficient) & 12 of 12 (circular) & One setting only & 0.025 (0.45) & 0.976 & 0.647 \\
\bottomrule
\end{tabularx}
\end{table}

\subsection*{The pigmentation-sweep response held across retraining, a second dataset and a null distribution}

\begin{figure}[tbp]
\centering
\includegraphics[width=\textwidth]{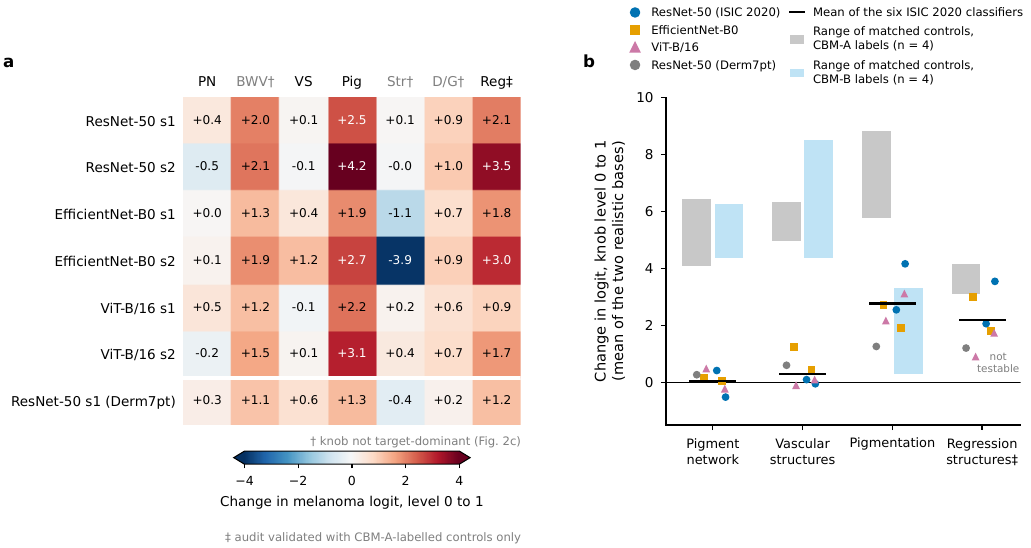}
\caption{\textbf{Counterfactual audit of the melanoma classifiers.} \textbf{a}, Change in the melanoma logit from knob level 0 to 1 (mean of the two realistic bases, 64 trajectories each) for the six ISIC 2020 classifiers and the Derm7pt classifier. \dag, knob not target-dominant (Fig.~\ref{fig:instrument}c). \textbf{b}, The four target-dominant knobs. Points, individual classifiers; black lines, mean of the six ISIC 2020 classifiers; grey and light-blue bars, range of the four matched positive controls with CBM-A and with CBM-B labels for the same knob; logit scales differ between classifiers, so the bars are a rough guide. \ddag, audit passed with CBM-A-labelled controls only.}
\label{fig:melanoma}
\end{figure}

On the realistic bases, the pigmentation sweep raised the melanoma logit of the six ISIC 2020 classifiers by +2.77 on average (per classifier +1.90 to +4.16) and the regression sweep by +2.17 (+0.90 to +3.54) (Fig.~\ref{fig:melanoma}). The vascular sweep gave +0.29 ($-0.10$ to +1.24; +0.01 without EfficientNet-B0) and the pigment-network sweep +0.05 ($-0.52$ to +0.48), with per-classifier 95\% intervals from $-1.67$ to +2.16 (Supplementary Table S7). The Derm7pt classifier showed the same order with smaller values (pigmentation +1.26, regression +1.20). Colour also moved the classifiers: changing only the colour target, with the same noise and concept vector, changed the logit by 0.84 to 2.49 on average, and darker images raised it ($-0.23$ to $-0.60$ logits per +10 $L^*$).

\begin{figure}[tbp]
\centering
\includegraphics[width=\textwidth]{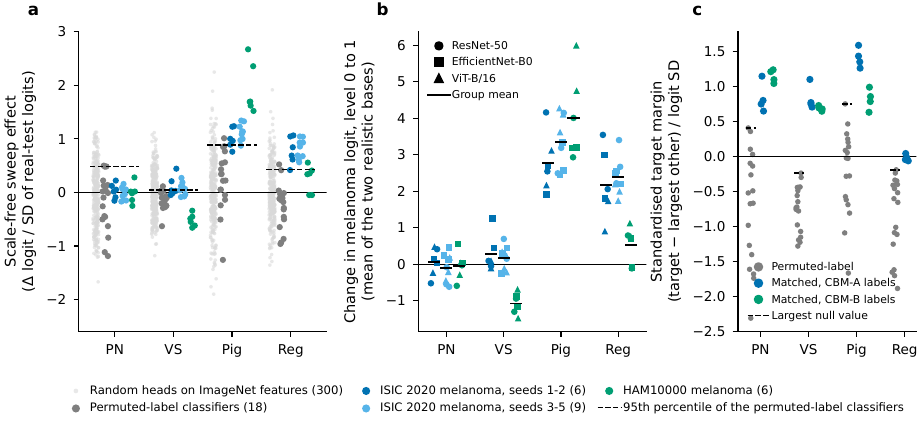}
\caption{\textbf{Null distribution, retraining and a second training dataset.} \textbf{a}, Scale-free effect of each target-dominant knob (logit change from level 0 to 1, mean of the two realistic bases, divided by the SD of the classifier's logits on the 969 real ISIC 2020 test images). Grey, null classifiers: 18 trained on permuted melanoma labels (dark) and 300 random linear heads on ImageNet features (light); dashed, 95th percentile of the permuted-label classifiers (rule 5). \textbf{b}, Logit effects of the 15 ISIC 2020 melanoma classifiers (seeds 1 and 2, dark blue; seeds 3 to 5, light blue; rule 6) and of the six HAM10000 classifiers (green; rule 7); black lines, group means. \textbf{c}, Standardised target margin (target effect minus the largest other effect, divided by the logit SD; the smaller of the two realistic bases) of the matched controls and of the permuted-label classifiers scored for the same target; dashed, largest null value. Post hoc. PN, pigment network; VS, vascular structures; Pig, pigmentation; Reg, regression structures.}
\label{fig:robust}
\end{figure}

These intervals cover generated trajectories for fixed models. Rules 5 to 7, fixed before their run, asked whether the findings depend on the seed, the training data, or a response that any image-sensitive model shows. With the nine retrained classifiers, the pigmentation effect had intervals above 0 on both bases in 15 of 15 ISIC 2020 classifiers (mean +3.12) and was the largest of the four target-dominant effects in 12 of 15 (Fig.~\ref{fig:robust}b). The spread between seeds of one architecture (mean SD 0.73 logits) was as large as the trajectory intervals (mean half-width 0.72; Supplementary Table S26). Six HAM10000 classifiers, audited with the same generator, responded to the pigmentation sweep in 6 of 6 (+4.01; +2.93 to +6.00; Supplementary Table S30), but not consistently to the regression sweep (+0.53; interval above 0 in 1 of 6), and the vascular sweep lowered their logit ($-1.08$).

The 18 permuted-label classifiers gave the null distribution. In scale-free units (effect divided by the SD of a classifier's logits on the real test images), the mean pigmentation effect of the six classifiers (1.01) and the mean regression effect (0.78) exceeded the 95th percentile of the null classifiers (0.88 and 0.43), each in five of six classifiers, and no architecture-matched group of six null classifiers reached either mean; the vascular effect exceeded it on average (0.11 against 0.05) but in two classifiers only, and pigment network did not (Fig.~\ref{fig:robust}a). With 18 null classifiers, the smallest tail probability is 0.053, so these are percentile checks, not tests at 0.05. Most null classifiers also responded most to the pigmentation sweep (11 of 18), but weakly (mean 0.23), so its rank alone says little. The null classifiers also exposed two weaknesses of our statistics: the per-classifier Wilcoxon test flagged 13\% of their knob-base cells as positive and significant, because it treats trajectories, not models, as the unit, and the rank-based control rule would have accepted 4 of 18 null classifiers as pigmentation controls and 1 as a pigment-network control. A standardised margin (target effect minus the largest other effect, divided by the logit SD) separated the matched controls from every null classifier for pigment network (8 of 8) and vascular structures (8 of 8) and for 7 of 8 pigmentation controls, whereas the regression controls had margins near 0 ($-0.06$ to +0.04; Fig.~\ref{fig:robust}c).

\subsection*{Knob cross-talk explains part of the disagreement between observation and intervention}

\begin{figure}[tbp]
\centering
\includegraphics[width=\textwidth]{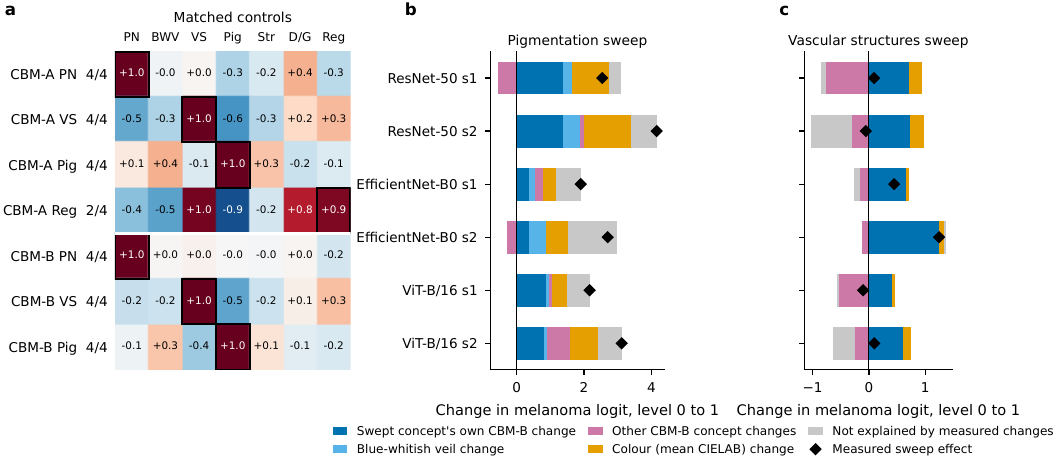}
\caption{\textbf{Decomposition of the sweep effects (post hoc).} A ridge regression relates the logit change of every image in the realistic sweeps to the changes in the seven CBM-B outputs and the mean CIELAB colour. \textbf{a}, Matched controls: mean CBM-B coefficient of the four controls of each group, scaled by the largest absolute coefficient of each control; outlined, target concept; n/4, controls with the target as largest coefficient and a bootstrap interval above 0. \textbf{b}, \textbf{c}, Parts of the pigmentation-sweep and vascular-sweep effects in the six ISIC 2020 classifiers: coefficient $\times$ mean measured change at level 1 (both realistic bases). Positive parts stack to the right and negative parts to the left of 0; grey, the part not explained by the measured changes; diamond, measured sweep effect.}
\label{fig:decomp}
\end{figure}

On real test images, vascular structures had the largest CBM-B coefficient for all six ISIC 2020 classifiers (+0.66 to +0.99 logits per SD), and the regression coefficient was negative for all six ($-0.07$ to $-0.51$); the pigment-network, vascular and pigmentation coefficients, but not the regression coefficient, exceeded the 95th percentile of the null classifiers. The audit gave the opposite picture. To find which image changes carried each sweep effect, a post hoc ridge regression related the logit change of every image in the realistic sweeps to the changes in the seven CBM-B outputs and in mean colour (Methods). Because images and predictors are shared, the controls test this design directly. The target had the largest coefficient, with an interval above 0, in 14 of 16 CBM-A controls (4 of 4 for pigment network, vascular structures and pigmentation; 2 of 4 for regression) and in all 12 CBM-B controls, the second test being partly circular (Fig.~\ref{fig:decomp}a).

In the melanoma classifiers, the measured changes explained 10\% to 29\% of the variance. The veil change carried only 3\% to 18\% of the pigmentation-sweep effect (+0.09 to +0.50 logits), against +0.38 to +1.38 for the pigmentation change and +0.38 to +1.40 for colour (Fig.~\ref{fig:decomp}b). On the 31 trajectories in which the pigmentation output rose more than every other CBM-B output, the sweep still raised the logit in all six classifiers (benign base +1.98 [+1.20, +2.83]; melanoma base +2.30 [+1.42, +3.22]). In the vascular sweep, the vascular change contributed +0.40 to +1.25 (intervals above 0 in 6 of 6, and +0.53 to +0.90 in 9 of 9 retrained classifiers), but the falls in other CBM-B outputs caused by the same sweep contributed $-0.12$ to $-0.77$ (Fig.~\ref{fig:decomp}c); in the HAM10000 classifiers the vascular part was negative. The ISIC 2020 classifiers therefore respond to vascular change, as observation suggests, and cross-talk hides part of this response in the sweep. The decomposition did not recover the regression controls, so the opposite signs for regression remain unexplained. An exploratory calibration of observational coefficients points the same way (Supplementary Fig.~S16).

\section*{Discussion}

This study asked whether a concept-level counterfactual audit can recover the training targets of control classifiers. It can, but only when the controls themselves are not confounded. With matched positive controls, the audit recovered the target in all 16 controls labelled by CBM-A and all 12 testable controls labelled by CBM-B, against 11 of 16 for our TCAV and 0 of 3 for Captum. One permuted-label control could not separate the methods, because output scale differs between classifiers; 18 of them, used as a null distribution, showed that significant per-classifier tests and target rankings also arise without melanoma labels, so audit effects need standardised margins against null classifiers. Observational attribution recovered the control targets and gave the most stable profiles, yet it described the melanoma classifiers differently from the audit; stability alone does not show validity.

The failure of the first test shows that a positive control can itself be confounded. A control built from quartiles of a sparse pseudo-label learned the concepts that travel with that label, and the audit reported them; TCAV passed the same control because its concept set came from the same confounded label. A positive control acts as a measurement standard, and a confounded standard can reject a method that recovers its target while it accepts a circular one.

The second set of matched controls reduces a circularity in our own design, in which the CBM-A pseudo-labels defined both the knobs and the first controls. It does not remove it, because CBM-B learned from the same Derm7pt images as CBM-A. Concept audits should therefore report balance in a second label set. A control also has a known training target, not a known mechanism: it may use features that travel with its target. Functions defined directly on an evaluator output gave the same verdicts as the knob test (Supplementary Table S28), so they test the knobs rather than the audit.

The matched controls also explain how observational attribution can pass a validity test and still disagree with an intervention. Matching reduces the correlation between the target concept and the other measured concepts, so for these controls association and intervention nearly coincide. The melanoma classifiers, by contrast, learned from ISIC 2020 images in which the concepts occur together. In that setting, a regression on concept scores divides the effect among correlated predictors, and the result depends on the correlation structure as much as on the classifier \cite{goyal2019cace,pearl2009causality}.

An intervention avoids this problem only if the knob changes one concept, and our knobs do not. The decomposition and the trajectory filter suggest that the pigmentation-sweep effect does not come mainly from the veil change, but they cannot exclude features that CBM-B does not measure, and most of the logit variance stayed unexplained. For vascular structures, cross-talk hid a response that both observation and the decomposition found. Each method alone therefore gave a partial picture. Expert ratings of each change and better disentangled knobs would be needed to isolate single concepts.

Our results agree with the main finding of DeGrave et al.\ that dermatology classifiers use lesional pigmentation patterns as well as colour \cite{degrave2025auditing}: here, the pigmentation-sweep response held in 15 ISIC 2020 classifiers with five seeds per architecture and in six HAM10000 classifiers, at about four times the mean null response, and colour alone moved the logit. The regression finding is the weakest: its controls had margins near 0, it rests on CBM-A labels only, and it did not transfer to HAM10000. Our audit works at the level of named concepts and is tested against controls with known training targets, but it lacks the dermatologist reading that DeGrave et al.\ used to name the features. The 7-point checklist scores an atypical pigment network and vascular pattern as major criteria and irregular blotches only as a minor criterion \cite{argenziano1998sevenpoint}, but we did not test with experts whether the generated images show atypical forms, so this contrast raises a question for clinical validation rather than showing a defect of the classifiers.

Part of the weak recovery by TCAV comes from the architecture: in the last layers of ResNet-50 and EfficientNet-B0, only average pooling and a linear head follow, so the gradient of the logit with respect to the layer output is the same for every image. Every TCAV run at these layers therefore gave a score of exactly 0 or 1, set only by the angle between the CAV and the head weights, which also explains the saturated Captum scores. Pattern-based CAVs \cite{pahde2025navigating} may behave differently. TCAV thus failed our criteria only in the implementations and layers studied.

This study has several limitations. First, the knobs are not isolated: three are not target-dominant, the four target-dominant knobs change other concepts, and all follow CBM-A pseudo-labels, which are model outputs. Both concept models learned from Derm7pt, where 30.3\% of the images share their seven-criterion profile with an image of a different diagnosis \cite{napoles2026derm7pt}. Second, the images are generated at $128\times128$ pixels, the melanoma prototype images are far from real images, and the colour calibration is weaker for the red channel. The classifiers separated generated melanoma and benign cases poorly (AUC 0.629 to 0.677), so the audit measures responses to concept sweeps, not diagnostic accuracy on generated images. Third, the generator comes from ISIC 2020, so the HAM10000 classifiers test transfer of the findings under the same instrument, not a second instrument; the generator saw all ISIC 2020 melanomas, including the classifier test melanomas; and the veil and regression pseudo-labels are sparse. Fourth, the trajectory intervals do not cover patients, and retraining added variation of similar size. Fifth, our TCAV is a pooled variant, Captum ran with default settings, and neither was run on the CBM-B controls or the later null classifiers. Sixth, the matched-control rules were fixed after the first rule had failed and none was registered externally; the decomposition, the trajectory filter, the standardised margins, the calibration and the scale-free analyses are post hoc, and with 18 null classifiers the null percentiles are imprecise.

Future work should test regression structures with a second label set on a larger dataset, add a dermatologist reading of the counterfactual images, use generators with higher resolution and better disentangled knobs, train the generator on other datasets, and use more null classifiers. The validation design itself transfers directly: any concept-level explanation method can be tested with matched positive controls on the classifiers it is meant to explain, before its output is used in the clinic.

\section*{Methods}

\subsection*{Data}

We used three public dermoscopy datasets for the instrument and two more for later tests. ISIC 2020 \cite{rotemberg2021isic} has 33,126 images, of which 584 are melanomas. Derm7pt \cite{kawahara2019derm7pt} has expert labels for the 7-point checklist concepts; we used its official split. ISIC 2018 Task 2 \cite{codella2019isic2018} has lesion-attribute masks; we used the presence of each mask as an image-level label. No ISIC 2018 image ID appears in ISIC 2020. HAM10000 \cite{tschandl2018ham10000} (10,015 images) trained a second set of melanoma classifiers, and PH2 \cite{mendonca2013ph2} (200 images: 80 common nevi, 80 atypical nevi and 40 melanomas, with expert dermoscopic criteria) tested the concept evaluators. All images were resized to $128\times128$ pixels for generation, scoring and classifier input. Classifiers and evaluators received the 128-pixel image upsampled to 224 pixels with ImageNet normalisation, so real and generated images followed the same pipeline. This study is a secondary analysis of publicly released, de-identified images, used under the terms of their providers \cite{rotemberg2021isic,kawahara2019derm7pt,codella2019isic2018,tschandl2018ham10000,mendonca2013ph2}. It involved no contact with patients and no new data collection, and we did not seek separate ethical approval.

The data had the following roles. The generator was trained on all 584 ISIC 2020 melanomas, each repeated five times, and on 8,760 benign images sampled at random without replacement (seed 42): 11,680 rows in total, with a melanoma-to-benign ratio of 1:3. The classifier data (below) therefore include melanomas that the generator saw in training, among them all 85 test melanomas. Of the 5,840 benign classifier images, 357 belong to a held-out set of 2,000 benign images that the generator did not see; the others were sampled independently of the generator's training sample, so some of them were also in generator training. The realistic bases are pseudo-label vectors of generator-training images: 64 benign images and 64 draws from 62 melanomas. The 584 melanomas and the held-out benign set served as real references for colour and realism. Classifier test AUCs are measured on real images and do not depend on the generator, but the audit as a whole is not an evaluation on images that the generator did not see. The code repository lists the known roles of each image.

\subsection*{Generator and colour-calibrated sampling}

The generator is a denoising diffusion probabilistic model \cite{ho2020ddpm} with a U-Net of five resolution levels (128 to 8 pixels; 64, 64, 128, 256 and 256 channels) and self-attention at the three lowest levels. Sinusoidal timestep embeddings pass through a two-layer MLP (256 dimensions, SiLU). The seven-dimensional concept vector passes through a separate two-layer MLP to a 128-dimensional embedding that is added in every residual block. The noise schedule is linear ($\beta$ from $1\times10^{-4}$ to 0.02, $T = 1{,}000$). The conditioning vectors are the CBM-A pseudo-labels of each ISIC 2020 training image. These rows were split at random into 10,512 training and 1,168 validation rows after the five-fold copying, so copies of one melanoma could fall on both sides; the validation loss was not used to choose the audit checkpoints. We trained for 150 epochs with AdamW \cite{loshchilov2019adamw} (learning rate $2\times10^{-4}$, weight decay $10^{-2}$), cosine annealing after a 10-epoch linear warm-up, gradient clipping at 1.0 and an effective batch size of 32 (micro-batches of 8 with four accumulation steps on GPU; 4 images on each of 8 cores on TPU). For classifier-free guidance \cite{ho2022cfg}, the conditioning vector was replaced by zeros with probability 0.15 during training. All audit images came from the exponential-moving-average weights (decay 0.999) at epoch 150. A baseline model with the same U-Net, data and training protocol is conditioned on a two-dimensional class vector only (15,989,891 parameters, against 15,990,531 for the concept-conditioned model). Training ran on NVIDIA T4 GPUs and a TPU v5e-8 on Kaggle.

We sampled with DDIM \cite{song2021ddim} (50 steps, leading timestep spacing) and classifier-free guidance at $s = 3.5$, with a zero vector as the unconditional input. Start noises were drawn on the CPU from fixed seeds. Images in one trajectory, and images that we compared directly, used the same noise.

Colour calibration works as follows. For each model, conditioning type and class, a linear map gives the mean RGB of the generated image from the three channel means of the start noise, $\mathbf{m}$: colour $= W\mathbf{m} + \mathbf{b}$. We fitted $W$ and $\mathbf{b}$ on pilot images at $s = 3.5$. For a colour target $\mathbf{t}$, we solved $W\mathbf{m}^* = \mathbf{t} - \mathbf{b}$ by least squares, replaced the channel means of the start noise by $\mathbf{m}^*$ and kept its spatial pattern. Colour targets were drawn from a Gaussian fitted to the per-image mean RGB of real images of the target class (584 melanomas; 2,000 benign images) and clipped to the observed range. Along a sweep trajectory, the colour target stayed constant. A class-agnostic control drew targets from an equal mixture of the two class Gaussians.

Colour fidelity used the per-image mean CIELAB (D65). An image was a colour outlier when its Mahalanobis distance to the pooled real images (melanoma and benign) exceeded the 99th percentile of the distances of those real images. Realism used FID \cite{heusel2017fid} and KID \cite{binkowski2018kid} with Inception features (KID on subsets of 500 images). The memorisation screen compared each melanoma-conditioned image with its nearest real training melanoma (cosine similarity of Inception features) and used the real leave-one-out distribution as reference. The ISIC 2020 release lists 425 pairs of duplicate images \cite{isic2020duplicates}, three of them melanomas, which is consistent with the largest real-to-real similarity of 1.000 in this reference.

\subsection*{Concept evaluators}

CBM-A is the concept bottleneck model \cite{koh2020cbm} that produced the pseudo-labels; on the Derm7pt test set it reached a mean concept AUC of 0.822. Because CBM-A made the conditioning labels, we did not use it to judge whether the knobs work. CBM-B is a ConvNeXt-Tiny \cite{liu2022convnext} with seven sigmoid outputs, trained on the Derm7pt expert labels (official split; AdamW \cite{loshchilov2019adamw}, learning rate $10^{-4}$, weight decay 0.05, one-cycle cosine schedule, batch 32, up to 40 epochs, early stopping on the mean validation AUC). The ISIC 2018 detector is a ResNet-50 \cite{he2016resnet} with five attribute outputs, trained in the same way on a random 70/15/15 split (seed 42). With the 128-pixel pipeline, CBM-B reached a mean test AUC of 0.849, and the detector reached 0.786 for pigment network. Both concept models use the same scoring of the Derm7pt labels: pigment network 0 (absent), 0.5 (typical) or 1 (atypical); pigmentation, streaks, and dots and globules 0 (absent), 0.5 (regular) or 1 (irregular); blue-whitish veil, vascular structures and regression structures 0 (absent) or 1 (present), with vascular structures within regression counted as absent. CBM-B was trained on these graded scores, and the evaluator AUCs count any value above 0 as present. Lesion darkness was 100 minus the mean $L^*$ inside the lesion, which we segmented by Otsu thresholding of $L^*$ in a central disk. We report evaluator changes in SD units of the 969 real classifier test images.

\subsection*{Concept sweeps}

A trajectory is one start noise, one colour target and one base vector, with one knob set to a series of levels. We used four bases. The two prototype bases are fixed class prototypes. For each concept, the melanoma prototype takes the 90th percentile of the pseudo-labels of the 584 ISIC 2020 melanomas, and the benign prototype takes the 10th percentile of the benign images. For blue-whitish veil, the melanoma value is the 98th percentile, because high veil values occur in only about 2\% of melanomas (13 of 584 reach 0.5). In the order pigment network, blue-whitish veil, vascular structures, pigmentation, streaks, dots and globules, regression structures, the melanoma prototype is (0.8292, 0.7781, 0.9926, 0.8918, 0.4585, 0.9256, 0.8769) and the benign prototype is (0.0800, 0.0002, 0.0047, 0.0055, 0.0018, 0.1501, 0.0003). Each prototype had 32 trajectories per knob at 11 levels (0 to 1 in steps of 0.1). The two realistic bases are pseudo-label vectors of generator-training images (64 benign images; 64 draws, with replacement, from 62 melanomas). Each realistic base had 64 trajectories per knob at 6 levels (0 to 1 in steps of 0.2). The other six concepts kept the values of the base. In total, the sweeps contain 4,928 prototype images and 5,376 realistic-base images. The melanoma prototype is extreme for some concepts: its blue-whitish-veil value of 0.7781 is higher than 99.91\% of the ISIC 2020 pseudo-labels (29 of 33,126 images reach it).

\subsection*{Target dominance of the knobs}

For each base, knob and evaluator output, we computed the mean change from level 0 to level 1 in SD units of real test images. A knob was target-dominant in a base when its own CBM-B output rose, and rose more than under any of the other six knobs. This criterion compares the knobs for one concept; it does not require the other outputs of a knob to stay unchanged, and we report them in Fig.~\ref{fig:instrument}c and Supplementary Tables S4 and S5. We did the same for the three ISIC 2018 outputs that correspond to a 7-point concept. We added this analysis after the audit runs. A generator check fixed in advance had required a positive, significant CBM-B dose response for pigmentation and pigment network on the melanoma prototype, confirmed by lesion darkness and by the ISIC 2018 detector.

\subsection*{Counterfactual audit statistics}

For each classifier, base and knob, we computed the melanoma logit of every image, clipped at $\pm15$. The effect is the mean over trajectories of the logit change from level 0 to level 1 ($\Delta$logit), with a 95\% bootstrap interval (2,000 resamples). We also report the mean change in $p(\text{melanoma})$. The test uses the ordinary least-squares slope of the logit on the level in each trajectory. A two-sided Wilcoxon signed-rank test compared the slopes with zero, and the matched-pairs rank-biserial correlation gives its effect size. We applied the Benjamini-Hochberg correction \cite{benjamini1995fdr} to the $P$ values of the seven knobs of one classifier and base. An effect was positive and significant when the corrected $p$ was below 0.05 and the median slope was above zero. The reported effects are mean endpoint differences, and their significance comes from the slopes. The intervals and tests describe variation across generated trajectories for a fixed generator, evaluators and classifier; they do not describe variation across patients or across retrained classifiers. Decision flips used the threshold that gives 90\% sensitivity for melanoma on the validation set.

The decomposition fitted a ridge regression ($\alpha = 1$, standardised predictors) of the logit change from level 0, at every level of every sweep, on the changes in the seven CBM-B outputs and in the mean $L^*$, $a^*$ and $b^*$. Intervals came from 200 bootstrap resamples of trajectories. The control test and the parts of each sweep effect (Fig.~\ref{fig:decomp}) used the two realistic bases (4,480 image pairs per classifier) and 500 and 300 resamples. A control passed when its target had the largest CBM-B coefficient and the lower bound of its interval was above 0. The part of a measured change in a knob's effect was its coefficient, on the original scale, times the mean change at level 1 of that knob; the rest was the measured effect minus all parts. Ordinary least squares, $\alpha = 10$ and a model without colour gave the same control results (26 of 28; 27 of 28 without colour) and pigmentation, veil and vascular parts within 0.4 logits (Supplementary Table S23). In the trajectory filter, the off-target change of a trajectory was the largest absolute change of the six other CBM-B outputs from level 0 to 1, in SD units of the real test images; we kept trajectories in which the target change exceeded it, or in which it was at most 1, 1.5 or 2 SD (Supplementary Table S24). Colour sensitivity used pairs of images from the colour-calibration experiment with the same noise and concept vector but a class-specific or a class-agnostic colour target. A linear regression related the logit change to the change in mean $L^*$, $a^*$ and $b^*$ (per 10 units).

\subsection*{Classifiers}

The classifier data contain all 584 ISIC 2020 melanomas and a random sample of 5,840 benign images (1:10). We split them by patient into 4,518 training, 937 validation and 969 test images (418, 81 and 85 melanomas). The architectures were ResNet-50 \cite{he2016resnet}, EfficientNet-B0 \cite{tan2019efficientnet} and ViT-B/16 \cite{dosovitskiy2021vit} from torchvision with ImageNet weights, followed by dropout (0.2) and one linear output. We trained with AdamW (learning rate $10^{-4}$ for the convolutional networks and $3\times10^{-5}$ for ViT-B/16, weight decay 0.05), a one-cycle cosine schedule with 10\% warm-up, batch size 32 and at most 15 epochs. Training stopped after 5 epochs without a rise in validation AUC, and we kept the best epoch. In the binary cross-entropy loss, we weighted the positive class by the ratio of negative to positive images, with a maximum of 20. Augmentation used a random 128-pixel crop of a 144-pixel image, horizontal and vertical flips, $90^\circ$ rotations and $\pm10\%$ brightness and contrast jitter. Training used mixed precision; every score in the audit used float32, with no non-finite values. Clipping of the logits at $\pm15$ affected 0.10\% or fewer of the images for 24 of the 26 classifiers, and 0.45\% and 4.38\% of the generated images for the two EfficientNet-B0 classifiers (lowest logit $-6{,}524.9$). We did not remove the duplicate images listed by ISIC \cite{isic2020duplicates}. All 425 listed pairs share a patient identifier, so the patient-level split kept each pair within one split; 22 pairs fall in the classifier data, and none crosses splits. The Derm7pt classifier used the same recipe on the Derm7pt training split (412 images, 90 melanomas).

\subsection*{Positive and negative controls}

The two quartile controls used the same architecture and recipe. Their label was 1 for the top quartile and 0 for the bottom quartile of one training-split pseudo-label (pigmentation or blue-whitish veil), and each had 2,260 training images. The negative control used melanoma labels permuted within the training split and within the validation split.

The matched controls compared high and low values of one pseudo-label. The low group was the bottom quartile of the training split. The high group was the top quartile, except for regression, where it was a pseudo-label of 0.5 or more, because 75\% of images have a regression pseudo-label below 0.004. We standardised the other six pseudo-labels on the training split. In a random order of the high images (seed 2029 plus the concept index), each high image took the nearest unused low image with the same melanoma label, by Euclidean distance on the six standardised labels. A pair was kept when the distance was at most 0.5 (1.0 for regression). We matched within each split separately. The low and high thresholds were 0.180 and 0.636 for pigment network, 0.012 and 0.603 for vascular structures, 0.016 and 0.232 for pigmentation, and 0.000 and 0.500 for regression. Balance is given as the standardised mean difference between the high and low members of the training pairs.

The matched controls above use CBM-A pseudo-labels, which also condition the generator. A control could therefore pass because it shares a bias of CBM-A with the knob, not because it detects the concept. To reduce this dependence, we built a second set of matched controls whose labels came from CBM-B, which did not make the conditioning labels (it was trained separately, on the same Derm7pt images as CBM-A). For each of the four target-dominant concepts, the low group was the bottom quartile of the CBM-B output on the training split. The high group was the top quartile, or a CBM-B output of 0.5 or more when the top quartile started below 0.5. We matched pairs 1:1 as above, but on the other six standardised CBM-B outputs, with a caliper of 0.5 (1.0 when the training split gave fewer than 100 pairs) and seed 3029 plus the concept index. A concept with fewer than 50 training pairs was not testable. Each concept had four controls (ResNet-50 seeds 1 to 3 and ViT-B/16 seed 1), trained with the same recipe. We report the balance of both sets of pairs in both label sets. As a check of reproducibility, ten classifiers of the earlier runs were scored again in the same run; all 280 of their effects were identical (largest difference 0.0).

\subsection*{Validity rules}

We wrote each decision rule into the analysis code and its configuration before the run that tested it. The rules were not registered with an external registry; the released code contains them as run.

\begin{enumerate}
\setlength{\itemsep}{1pt}
\item Original audit rule (before the first audit run): each quartile positive control shows its target as the largest positive and significant effect in at least 3 of 4 bases, and the largest absolute effect of the negative control is below 0.25 times the median largest absolute effect of the melanoma classifiers.
\item Matched-control rule (before the run that trained the first three matched controls): a matched control passes when its target knob has the largest positive and significant effect in at least 3 of 4 bases.
\item Per-concept rule (before the run that added 13 matched controls): the audit is valid for a concept when at least 3 of its 4 matched controls pass. Earlier verdicts were not changed.
\item Second-label rule (written into the code and printed with its SHA-256 fingerprint at the start of the run, before the run computed the CBM-B labels and the pairs): a CBM-B matched control passes when its target knob has the largest positive and significant effect in at least 3 of 4 bases. The audit is valid for a concept when at least 3 of its 4 CBM-B matched controls pass, and a concept that is not testable is not valid. We claim validity only for the concepts that pass this rule.
\item Rules 5 to 8 (written into the code of the third run and printed with their SHA-256 fingerprint before the run trained any classifier). Rule 5: a melanoma finding for a knob is above null when the mean scale-free effect of the six primary melanoma classifiers is above the 95th percentile of the 18 permuted-label classifiers. Rule 6: the pigmentation finding is stable under retraining when its interval is above 0 on both realistic bases in at least 12 of 15 ISIC 2020 classifiers and pigmentation has the largest mean effect of the four target-dominant knobs. Rule 7: the pigmentation finding transfers to HAM10000 when its interval is above 0 on both realistic bases in at least 5 of 6 HAM10000 classifiers and pigmentation has the largest mean effect of the four knobs. Rule 8 (descriptive): CBM-B and the ISIC 2018 detector are compared with the PH2 expert criteria by AUC.
\end{enumerate}

Rules 2 and 3 were defined after rule 1 failed. They therefore test a design that we chose with knowledge of the first failure, although not with knowledge of their own outcome. Rule 4 was defined after rules 2 and 3 had passed, in order to remove the dependence of the controls on CBM-A, and before its own outcome was known. The fingerprint of rule 4 is \texttt{\seqsplit{919a511cf55ae6f5b249ba11c465ea25a460529fb775e49913c4d5a360ccf880}}, and the released code prints the same value. A fingerprint identifies the text of a rule; on its own, it does not prove when the text was written. The limit of 0.25 in rule 1 was a pragmatic choice made before the first run, not a value derived from theory. The fingerprint of rules 5 to 8 is \texttt{\seqsplit{287492fb0ca91d0dbaee3f839345472d28fec594ac208c02e6cdf87dad20e318}}.

\subsection*{Output scale of the negative control}

Rule 1 compares logit changes, whose size depends on the spread of a classifier's logits. In a post hoc analysis, we divided each classifier's effects by the SD of its logits on the 969 real test images (clipped at $\pm15$) and recomputed the ratio of rule 1 for the audit and for the observational attribution, with the same seven melanoma classifiers. As alternative measures of scale, we used the interquartile range of the test logits divided by 1.349 and the SD of the logits on the realistic-sweep images. TCAV scores do not depend on output scale and were not changed. Supplementary Table S22 gives all values.

\subsection*{Retraining, a second training dataset and null classifiers}

The third run used the same classifier data, recipe and audit images as Phase 2. It added nine ISIC 2020 melanoma classifiers (ResNet-50, EfficientNet-B0 and ViT-B/16, seeds 3 to 5) and 17 permuted-label classifiers (ResNet-50 seeds 2 to 10, EfficientNet-B0 seeds 1 to 5, ViT-B/16 seeds 1 to 3); each seed permuted the training and validation labels differently. The HAM10000 classifiers \cite{tschandl2018ham10000} learned melanoma against all other diagnoses from 10,015 images (1,113 melanomas) of 7,470 lesions. We split them by lesion (stratified grouped 20-fold partition, seed 2027) into 7,040 training, 1,530 validation and 1,445 test images; no image name occurred in ISIC 2020. They reached a test AUC of 0.911 to 0.929 on HAM10000 and 0.652 to 0.776 on the ISIC 2020 test images, and the same ResNet-50, EfficientNet-B0 and ViT-B/16 backbones were trained with seeds 1 and 2. A follow-up run, with checks fixed before it, rebuilt the same split and found no 128-pixel HAM10000 image identical to one of the 6,424 ISIC 2020 classifier images, no identical images under different lesion IDs and no lesion in two splits (Supplementary Table S32). A near-duplicate screen (cosine similarity of ImageNet ResNet-50 features; report only) found no HAM10000-ISIC 2020 pair at 0.98 or above (191 at 0.95; the 12 closest pairs showed different lesions on visual review, Supplementary Fig.~S18) and one pair of HAM10000 images from different lesions at 0.98, in the training and validation splits. These checks covered the ISIC 2020 classifier images, not all 33,126 ISIC 2020 images.

For rule 8, a mapping fixed before the analysis (Supplementary Table S31) compared the CBM-B outputs with the PH2 criteria: atypical against typical pigment network; dots and globules present, and atypical; streaks, regression areas and blue-whitish veil present. PH2 does not score vascular structures or pigmentation blotches, so these concepts were not testable. We report the AUC with a 95\% bootstrap interval (2,000 resamples) and the numbers of positive and negative images. The first run could not read the PH2 annotation file, whose six criteria are separated by single bars; the follow-up run used a corrected parser and no other change. On PH2, the six primary melanoma classifiers separated melanomas from nevi with AUCs of 0.747 to 0.961. As a second, untrained null, 100 random linear heads per architecture (Gaussian weights scaled by $d^{-1/2}$, fixed seed) read the centred penultimate ImageNet features. All classifiers were scored in float32 on the realistic and prototype sweeps and the real test images.

The scale-free effect divides a classifier's effect by the SD of its logits on the 969 real ISIC 2020 test images and averages the two realistic bases. For each melanoma classifier and knob, the empirical upper-tail probability was $(1 + \#\{\text{null} \geq T\})/(R + 1)$ against the $R = 18$ permuted-label classifiers (smallest value 0.053) and against the 300 random heads (Supplementary Table S27). The false-positive rate of the per-classifier test is the share of null-classifier, base and knob cells in which the Wilcoxon test of rule 2 was positive and significant. The standardised target margin (post hoc) is the target effect minus the largest effect of the six other knobs, divided by the logit SD; we report the smaller value of the two realistic bases with a paired bootstrap over base indices (2,000 resamples), and we scored each null classifier for every target (Supplementary Table S29). Between-architecture, between-seed and trajectory variation are reported separately (Supplementary Table S26). As known-function controls (post hoc), fixed functions of one evaluator output (each CBM-B concept, the ISIC 2018 pigment-network output, lesion darkness and mean $b^*$), standardised on the real test images, were audited in the same way (Supplementary Table S28).

\subsection*{TCAV, our implementation}

We used two layers per architecture: layer3 and layer4 of ResNet-50, blocks 5 and 8 of the EfficientNet-B0 feature stack, and encoder blocks 6 and 12 of ViT-B/16. Activations were averaged over spatial positions or tokens, and the gradient of the melanoma logit with respect to the layer output was summed over the same positions. A concept activation vector (CAV) \cite{kim2018tcav} was the normalised weight vector of an L2-regularised logistic regression ($C = 1$) that separated a concept set from a random set; 25\% of each set was held out to measure CAV accuracy. The TCAV score was the share of target images with a positive directional derivative. The target images were all 166 validation and test melanomas and 166 benign images from the same splits.

There were three concept-set types. For the two ISIC types, the pool was the top decile of the training images for one pseudo-label, either among all training images (452 images) or among benign training images only (410 images); each concept set had 100 images. For the Derm7pt type, the pool was the images with the concept marked present by the experts (142 to 448 images), and each set had 60 images. Random sets came from the rest of the same pool. Each concept had 20 runs with new concept and random sets, and the null distribution came from 20 runs that compared one fixed random set with other random sets. A Welch $t$-test compared the concept scores with the null scores, with Benjamini-Hochberg correction within the seven concepts. Ties for the highest mean score were broken by the mean cosine between gradient and CAV. The TCAV effect was the mean score minus the null mean. The settings for the rule were 2 layers $\times$ 2 ISIC set types $\times$ 2 target sets; the Derm7pt sets are reported but were not part of the rule. In the last layers of ResNet-50 and EfficientNet-B0, only global average pooling and the linear head follow, so the summed gradient equals the head weight vector for every image; each run score at these layers is therefore 0 or 1. Computed separately for each layer, the negative-control ratio was 0.65 for the middle layers and 0.90 for the last layers (0.834 for both together).

\subsection*{TCAV, Captum}

We used the TCAV class of Captum 0.9.0 \cite{kokhlikyan2020captum} with its default settings: the default linear classifier on flattened activations, sign-count scores, and the last of our two layers (the output of encoder block 12 for ViT-B/16). Each concept set was paired with 10 random sets. The two ISIC set types and the two target sets gave 4 settings, and the rules required 3 of 4. We ran Captum on 13 classifiers: the 7 melanoma classifiers, the 3 original controls and the first 3 matched controls. Captum gives no cosine, so ties were broken by concept order. This choice affected 2 settings of the pigmentation quartile control; if a tie counts as highest, that control passes in 4 of 4 settings, but the verdict does not change because the negative control fails. Captum reported CAV accuracies centred on 0.5 for concept and random CAVs alike (mean 0.500, range 0.35 to 0.67 over 2,080 CAVs), and it also reported 0.5 on a toy example with two perfectly separable sets. In version 0.9.0, the library takes the arg-max over a one-column output for two-class CAVs, so the reported value tracks the class shares of the held-out data rather than the separation accuracy. The current Captum code handles this case. The CAV weights, and so the TCAV scores, are not affected.

\subsection*{Observational attribution and calibration}

The observational attribution fitted a ridge regression ($\alpha = 1$) of each classifier's melanoma logit on the seven CBM-B outputs and the mean $L^*$, $a^*$ and $b^*$ of the 969 real test images. Predictors were standardised, so a coefficient is the logit change per SD. For the exploratory calibration, defined after the audit results, we divided each matched control's audit effect for its target (mean of the two realistic bases) by its observational coefficient for the same concept. The median over the four controls of a concept gave a ratio, and a melanoma classifier's observational coefficient times this ratio gave the predicted audit effect.

\subsection*{Reliability and agreement}

For the audit, a classifier's profile was the vector of its 28 effects (7 knobs $\times$ 4 bases). For TCAV, it was the vector of TCAV effects over all settings, and for the observational method, the seven CBM-B coefficients. Reliability was the mean Spearman $\rho$ between profiles of classifier pairs: pairs of the same architecture with another seed, and pairs of different classifiers.

\subsection*{Software and hardware}

We used Python with PyTorch 2.11.0 (CUDA 12.8), torchvision, diffusers (DDIM scheduler), scikit-learn, SciPy and Captum 0.9.0. All audit runs used two NVIDIA Tesla T4 GPUs on Kaggle. A later run rebuilt the classifier data and confirmed that the images, split and pseudo-labels were identical; it also reproduced the earlier TCAV scores exactly (12,480 runs, largest difference 0.0).

\subsection*{Use of a large language model}

During this work, the authors used Claude (Anthropic) to assist with code development and debugging, to discuss analysis options, and for writing support and language editing of the manuscript text. The authors decided on the study design and on every decision rule before the run that tested it, ran all experiments, verified each reported result against the output files, and revised the full manuscript. The authors take full responsibility for the content of this article.

\section*{Data availability}

The source datasets are public. ISIC 2020 is described in ref.~\citenum{rotemberg2021isic} (\href{https://doi.org/10.1038/s41597-021-00815-z}{doi:10.1038/s41597-021-00815-z}). Derm7pt is available from its authors (\url{https://derm.cs.sfu.ca}). ISIC 2018 Task 2 is available from the ISIC challenge site (\url{https://challenge.isic-archive.com/landing/2018/}). HAM10000 is described in ref.~\citenum{tschandl2018ham10000} (\href{https://doi.org/10.1038/sdata.2018.161}{doi:10.1038/sdata.2018.161}). PH2 is available from its providers after registration (\url{https://www.fc.up.pt/addi/ph2\%20database.html}); its terms do not allow redistribution, so we share only our derived scores. The result tables behind every figure and table, the role of each image in generator training, the classifier splits and the audit bases, and the supplementary files are in the repository named below. The trained checkpoints (CBM-A, the two diffusion models, CBM-B, the ISIC 2018 detector and the classifiers), the CBM-A pseudo-labels, the generated images and the per-image scores are not in the repository because of their size; they are available from the corresponding author on request, including to referees during review.

\section*{Code availability}

All custom code is central to the conclusions and is available under the MIT licence at \url{https://github.com/srahman7900/melanoma-concept-audit}, with no restrictions on use. The repository holds the notebooks that trained CBM-A, wrote the pseudo-labels and trained the two diffusion models; the audit cells, which contain each decision rule as it was run; the offline scripts that rebuild every main figure and the post hoc tables from the released outputs, together with the inputs those scripts read; and the decision rules with their fingerprints.


\begin{thebibliography}{39}
\providecommand{\natexlab}[1]{#1}
\providecommand{\url}[1]{\texttt{#1}}
\expandafter\ifx\csname urlstyle\endcsname\relax
  \providecommand{\doi}[1]{doi: #1}\else
  \providecommand{\doi}{doi: \begingroup \urlstyle{rm}\Url}\fi

\bibitem[Winkler et~al.(2019)]{winkler2019markings}
J.~K. Winkler et~al.
\newblock Association between surgical skin markings in dermoscopic images and diagnostic performance of a deep learning convolutional neural network for melanoma recognition.
\newblock \emph{JAMA Dermatol.}, 155:\penalty0 1135-1141, 2019.
\newblock \doi{10.1001/jamadermatol.2019.1735}.

\bibitem[Bissoto et~al.(2020)Bissoto, Valle, and Avila]{bissoto2020debiasing}
A.~Bissoto, E.~Valle, and S.~Avila.
\newblock Debiasing skin lesion datasets and models? {N}ot so fast.
\newblock In \emph{Proc. IEEE/CVF Conference on Computer Vision and Pattern Recognition Workshops}, pages 3192-3201, 2020.

\bibitem[DeGrave et~al.(2025)DeGrave, Cai, Janizek, Daneshjou, and Lee]{degrave2025auditing}
A.~J. DeGrave, Z.~R. Cai, J.~D. Janizek, R.~Daneshjou, and S.-I. Lee.
\newblock Auditing the inference processes of medical-image classifiers by leveraging generative {AI} and the expertise of physicians.
\newblock \emph{Nat. Biomed. Eng.}, 9:\penalty0 294-306, 2025.
\newblock \doi{10.1038/s41551-023-01160-9}.

\bibitem[Kim et~al.(2018)]{kim2018tcav}
B.~Kim et~al.
\newblock Interpretability beyond feature attribution: quantitative testing with concept activation vectors ({TCAV}).
\newblock In \emph{Proc. 35th International Conference on Machine Learning}, volume~80 of \emph{PMLR}, pages 2668-2677, 2018.

\bibitem[Lucieri et~al.(2020)]{lucieri2020cav}
A.~Lucieri et~al.
\newblock On interpretability of deep learning based skin lesion classifiers using concept activation vectors.
\newblock In \emph{Proc. International Joint Conference on Neural Networks ({IJCNN})}. IEEE, 2020.
\newblock Preprint at arXiv:2005.02000.

\bibitem[Lucieri et~al.(2022)]{lucieri2022exaid}
A.~Lucieri et~al.
\newblock {ExAID}: a multimodal explanation framework for computer-aided diagnosis of skin lesions.
\newblock \emph{Comput. Methods Programs Biomed.}, 215:\penalty0 106620, 2022.
\newblock \doi{10.1016/j.cmpb.2022.106620}.

\bibitem[Argenziano et~al.(1998)]{argenziano1998sevenpoint}
G.~Argenziano et~al.
\newblock Epiluminescence microscopy for the diagnosis of doubtful melanocytic skin lesions. {C}omparison of the {ABCD} rule of dermatoscopy and a new 7-point checklist based on pattern analysis.
\newblock \emph{Arch. Dermatol.}, 134:\penalty0 1563-1570, 1998.
\newblock \doi{10.1001/archderm.134.12.1563}.

\bibitem[Kawahara et~al.(2019)Kawahara, Daneshvar, Argenziano, and Hamarneh]{kawahara2019derm7pt}
J.~Kawahara, S.~Daneshvar, G.~Argenziano, and G.~Hamarneh.
\newblock Seven-point checklist and skin lesion classification using multitask multimodal neural nets.
\newblock \emph{IEEE J. Biomed. Health Inform.}, 23:\penalty0 538-546, 2019.
\newblock \doi{10.1109/JBHI.2018.2824327}.

\bibitem[Adebayo et~al.(2018)]{adebayo2018sanity}
J.~Adebayo et~al.
\newblock Sanity checks for saliency maps.
\newblock In \emph{Advances in Neural Information Processing Systems}, volume~31, pages 9505-9515, 2018.

\bibitem[Ramaswamy et~al.(2023)Ramaswamy, Kim, Fong, and Russakovsky]{ramaswamy2023overlooked}
V.~V. Ramaswamy, S.~S.~Y. Kim, R.~Fong, and O.~Russakovsky.
\newblock Overlooked factors in concept-based explanations: dataset choice, concept learnability, and human capability.
\newblock In \emph{Proc. IEEE/CVF Conference on Computer Vision and Pattern Recognition}, pages 10932-10941, 2023.

\bibitem[Adebayo et~al.(2022)Adebayo, Muelly, Abelson, and Kim]{adebayo2022posthoc}
J.~Adebayo, M.~Muelly, H.~Abelson, and B.~Kim.
\newblock Post hoc explanations may be ineffective for detecting unknown spurious correlation.
\newblock In \emph{International Conference on Learning Representations}, 2022.
\newblock Preprint at arXiv:2212.04629.

\bibitem[Pahde et~al.(2025)]{pahde2025navigating}
F.~Pahde et~al.
\newblock Navigating neural space: revisiting concept activation vectors to overcome directional divergence.
\newblock In \emph{International Conference on Learning Representations}, 2025.
\newblock Preprint at arXiv:2202.03482.

\bibitem[Goyal et~al.(2019)Goyal, Feder, Shalit, and Kim]{goyal2019cace}
Y.~Goyal, A.~Feder, U.~Shalit, and B.~Kim.
\newblock Explaining classifiers with causal concept effect ({CaCE}), 2019.
\newblock Preprint at arXiv:1907.07165.

\bibitem[Pearl(2009)]{pearl2009causality}
J.~Pearl.
\newblock \emph{Causality: Models, Reasoning, and Inference}.
\newblock Cambridge Univ. Press, 2nd edition, 2009.

\bibitem[Abraham et~al.(2022)]{abraham2022cebab}
E.~D. Abraham et~al.
\newblock {CEBaB}: estimating the causal effects of real-world concepts on {NLP} model behavior.
\newblock In \emph{Advances in Neural Information Processing Systems}, volume~35, 2022.
\newblock Preprint at arXiv:2205.14140.

\bibitem[Varshney et~al.(2025)Varshney, Lucieri, Balada, Dengel, and Ahmed]{varshney2024counterfactual}
P.~Varshney, A.~Lucieri, C.~P. Balada, A.~Dengel, and S.~Ahmed.
\newblock Generating counterfactual trajectories with latent diffusion models for concept discovery.
\newblock In \emph{Pattern Recognition. ICPR 2024}, volume 15312 of \emph{Lecture Notes in Computer Science}, pages 138-153. Springer, 2025.
\newblock \doi{10.1007/978-3-031-78198-8_10}.

\bibitem[Kim et~al.(2024)]{kim2024monet}
C.~Kim et~al.
\newblock Transparent medical image {AI} via an image-text foundation model grounded in medical literature.
\newblock \emph{Nat. Med.}, 30:\penalty0 1154-1165, 2024.
\newblock \doi{10.1038/s41591-024-02887-x}.

\bibitem[Fayyad et~al.(2026)Fayyad, Bayasi, Yu, and Najjaran]{fayyad2025lesiongen}
J.~Fayyad, N.~Bayasi, Z.~Yu, and H.~Najjaran.
\newblock {LesionGen}: a concept-guided diffusion model for dermatology image synthesis.
\newblock In \emph{Skin Image Analysis, and Computer-Aided Pelvic Imaging for Female Health. ISIC 2025, CAPI 2025}, volume 16149 of \emph{Lecture Notes in Computer Science}. Springer, 2026.
\newblock \doi{10.1007/978-3-032-05825-6_1}.

\bibitem[Ho et~al.(2020)Ho, Jain, and Abbeel]{ho2020ddpm}
J.~Ho, A.~Jain, and P.~Abbeel.
\newblock Denoising diffusion probabilistic models.
\newblock In \emph{Advances in Neural Information Processing Systems}, volume~33, pages 6840-6851, 2020.

\bibitem[Ho and Salimans(2022)]{ho2022cfg}
J.~Ho and T.~Salimans.
\newblock Classifier-free diffusion guidance, 2022.
\newblock Preprint at arXiv:2207.12598; presented at a NeurIPS 2021 workshop.

\bibitem[Rotemberg et~al.(2021)]{rotemberg2021isic}
V.~Rotemberg et~al.
\newblock A patient-centric dataset of images and metadata for identifying melanomas using clinical context.
\newblock \emph{Sci. Data}, 8:\penalty0 34, 2021.
\newblock \doi{10.1038/s41597-021-00815-z}.

\bibitem[Koh et~al.(2020)]{koh2020cbm}
P.~W. Koh et~al.
\newblock Concept bottleneck models.
\newblock In \emph{Proc. 37th International Conference on Machine Learning}, volume 119 of \emph{PMLR}, pages 5338-5348, 2020.

\bibitem[Codella et~al.(2019)]{codella2019isic2018}
N.~Codella et~al.
\newblock Skin lesion analysis toward melanoma detection 2018: a challenge hosted by the {International Skin Imaging Collaboration (ISIC)}, 2019.
\newblock Preprint at arXiv:1902.03368.

\bibitem[Mendon{\c{c}}a et~al.(2013)Mendon{\c{c}}a, Ferreira, Marques, Mar{\c{c}}al, and Rozeira]{mendonca2013ph2}
Teresa Mendon{\c{c}}a, Pedro~M. Ferreira, Jorge~S. Marques, Andr{\'e} R.~S. Mar{\c{c}}al, and Jorge Rozeira.
\newblock {PH}$^2$ - a dermoscopic image database for research and benchmarking.
\newblock In \emph{35th Annual International Conference of the IEEE Engineering in Medicine and Biology Society (EMBC)}, pages 5437--5440, 2013.
\newblock \doi{10.1109/EMBC.2013.6610779}.

\bibitem[Lin et~al.(2024)Lin, Liu, Li, and Yang]{lin2024schedules}
S.~Lin, B.~Liu, J.~Li, and X.~Yang.
\newblock Common diffusion noise schedules and sample steps are flawed.
\newblock In \emph{Proc. IEEE/CVF Winter Conference on Applications of Computer Vision}, pages 5404-5411, 2024.

\bibitem[Everaert et~al.(2024)]{everaert2024signalleak}
M.~N. Everaert et~al.
\newblock Exploiting the signal-leak bias in diffusion models.
\newblock In \emph{Proc. IEEE/CVF Winter Conference on Applications of Computer Vision}, pages 4025-4034, 2024.

\bibitem[Tschandl et~al.(2018)Tschandl, Rosendahl, and Kittler]{tschandl2018ham10000}
Philipp Tschandl, Cliff Rosendahl, and Harald Kittler.
\newblock The {HAM10000} dataset, a large collection of multi-source dermatoscopic images of common pigmented skin lesions.
\newblock \emph{Scientific Data}, 5:\penalty0 180161, 2018.
\newblock \doi{10.1038/sdata.2018.161}.

\bibitem[Kokhlikyan et~al.(2020)]{kokhlikyan2020captum}
N.~Kokhlikyan et~al.
\newblock Captum: a unified and generic model interpretability library for {PyTorch}, 2020.
\newblock Preprint at arXiv:2009.07896.

\bibitem[N{\'a}poles et~al.(2026)N{\'a}poles, Grau, and Salgueiro]{napoles2026derm7pt}
G.~N{\'a}poles, I.~Grau, and Y.~Salgueiro.
\newblock Concept inconsistency in dermoscopic concept bottleneck models: a rough-set analysis of the {Derm7pt} dataset.
\newblock \emph{Sci. Rep.}, 16:\penalty0 26545, 2026.
\newblock \doi{10.1038/s41598-026-56927-2}.

\bibitem[Loshchilov and Hutter(2019)]{loshchilov2019adamw}
I.~Loshchilov and F.~Hutter.
\newblock Decoupled weight decay regularization.
\newblock In \emph{International Conference on Learning Representations}, 2019.

\bibitem[Song et~al.(2021)Song, Meng, and Ermon]{song2021ddim}
J.~Song, C.~Meng, and S.~Ermon.
\newblock Denoising diffusion implicit models.
\newblock In \emph{International Conference on Learning Representations}, 2021.

\bibitem[Heusel et~al.(2017)Heusel, Ramsauer, Unterthiner, Nessler, and Hochreiter]{heusel2017fid}
M.~Heusel, H.~Ramsauer, T.~Unterthiner, B.~Nessler, and S.~Hochreiter.
\newblock {GANs} trained by a two time-scale update rule converge to a local {Nash} equilibrium.
\newblock In \emph{Advances in Neural Information Processing Systems}, volume~30, 2017.

\bibitem[Bi{\'n}kowski et~al.(2018)Bi{\'n}kowski, Sutherland, Arbel, and Gretton]{binkowski2018kid}
M.~Bi{\'n}kowski, D.~J. Sutherland, M.~Arbel, and A.~Gretton.
\newblock Demystifying {MMD} {GANs}.
\newblock In \emph{International Conference on Learning Representations}, 2018.

\bibitem[{International Skin Imaging Collaboration}(2020)]{isic2020duplicates}
{International Skin Imaging Collaboration}.
\newblock {ISIC} 2020 challenge data: list of duplicate images in the training set ({ISIC\_2020\_Training\_Duplicates.csv}).
\newblock \url{https://challenge.isic-archive.com/data}, 2020.
\newblock Accessed 20 October 2026.

\bibitem[Liu et~al.(2022)]{liu2022convnext}
Z.~Liu et~al.
\newblock A {ConvNet} for the 2020s.
\newblock In \emph{Proc. IEEE/CVF Conference on Computer Vision and Pattern Recognition}, pages 11976-11986, 2022.

\bibitem[He et~al.(2016)He, Zhang, Ren, and Sun]{he2016resnet}
K.~He, X.~Zhang, S.~Ren, and J.~Sun.
\newblock Deep residual learning for image recognition.
\newblock In \emph{Proc. IEEE Conference on Computer Vision and Pattern Recognition}, pages 770-778, 2016.

\bibitem[Benjamini and Hochberg(1995)]{benjamini1995fdr}
Y.~Benjamini and Y.~Hochberg.
\newblock Controlling the false discovery rate: a practical and powerful approach to multiple testing.
\newblock \emph{J. R. Stat. Soc. B}, 57:\penalty0 289-300, 1995.
\newblock \doi{10.1111/j.2517-6161.1995.tb02031.x}.

\bibitem[Tan and Le(2019)]{tan2019efficientnet}
M.~Tan and Q.~Le.
\newblock {EfficientNet}: rethinking model scaling for convolutional neural networks.
\newblock In \emph{Proc. 36th International Conference on Machine Learning}, volume~97 of \emph{PMLR}, pages 6105-6114, 2019.

\bibitem[Dosovitskiy et~al.(2021)]{dosovitskiy2021vit}
A.~Dosovitskiy et~al.
\newblock An image is worth 16x16 words: transformers for image recognition at scale.
\newblock In \emph{International Conference on Learning Representations}, 2021.

\end{thebibliography}

\section*{Author contributions}

M.S.R.: conceptualization, methodology, software, validation, formal analysis, investigation, data curation, project administration and writing (original draft). D.N., M.K.M. and T.H.L.: supervision and writing (review and editing). M.H.I.: writing (review and editing), visualization. All authors read and approved the submitted version and agree to be accountable for their contributions.

\section*{Funding}

This research received no specific grant from any funding agency in the public, commercial or not-for-profit sectors.

\section*{Declarations}

\noindent\textbf{Competing interests.} The authors declare no competing interests.

\noindent\textbf{Ethics approval and consent to participate.} Not applicable. The study used only publicly released, de-identified dermoscopy images (ISIC 2020, Derm7pt, ISIC 2018 Task 2, HAM10000 and PH2) under the terms set by their providers. The authors collected no new data from people and had no contact with patients, so no separate ethics approval or informed consent was sought.

\section*{Additional information}

\noindent\textbf{Supplementary information} accompanies this paper. Supplementary Figures S1 to S18 are in the Supplementary Information file, and Supplementary Tables S1 to S32 are in Supplementary Data 1.

\noindent\textbf{Correspondence} and requests for materials should be addressed to T.H.L.

\end{document}